\PassOptionsToPackage{unicode}{hyperref}
\PassOptionsToPackage{hyphens}{url}
\documentclass[]{article}

\usepackage[T1]{fontenc}
\usepackage[utf8]{inputenc}
\usepackage{textcomp}
\usepackage{lmodern}

\usepackage{amsmath,amssymb}

\usepackage[margin=1in]{geometry}
\usepackage{parskip}

\usepackage{tabularx}
\usepackage{booktabs}
\usepackage{multirow}
\usepackage{array}
\usepackage{longtable}

\usepackage{graphicx}
\usepackage{float}

\usepackage{natbib}

\usepackage{xcolor}
\usepackage{microtype}
\usepackage{xurl}
\usepackage{setspace}

\usepackage{hyperref}
\hypersetup{hidelinks, pdfcreator={LaTeX via pandoc}}
\usepackage{cleveref}

\title{Closing the Modality Gap in Zero-Shot HAR: Contrastive Training and Separability-Optimized Prototypes on IMU Data}
\author{Anik Ghosh \\ avickg@gmail.com}
\date{}

\begin{document}

\maketitle

% \begin{abstract}
% Zero-Shot Learning (ZSL) for Inertial Measurement Unit (IMU)-based Human
% Activity Recognition (HAR) faces a key challenge: bridging the gap
% between sensor data and semantic class representations. Here, we
% systematically evaluate seven methods using the PAMAP2 dataset, with 14
% seen and 4 unseen activity classes, and hold subjects 108 and 109
% for testing. We find that the modality gap arises mainly during
% training. For example, a typical TCN model trained with contrastive
% cross-entropy over label-name SBERT prototypes yields sensor embeddings
% averaging a 0.30 cosine similarity with rich-description text
% prototypes, while switching to a contrastive loss with Sentence-BERT
% prototypes boosts this to 0.69. This adjustment consistently improves
% all three inference methods tested. The strongest results come from
% combining contrastive training with Inverted Softmax correction which
% achieves 73.2\% accuracy and a 0.583 macro F1 in unseen classes,
% compared to 58.3\% and 0.34 for the baseline. Our findings also
% highlight that overall accuracy can be misleading for imbalanced test
% sets and recommend macro-averaged F1 as the benchmark metric. The code
% is available at
% \url{https://github.com/AnikGhosh332/Human-Activity-Recognition-Using-Sensor-Data}.

% \vspace{1em}
% \noindent\textbf{Keywords:} Zero-Shot Learning, Human Activity Recognition,
% Inertial Measurement Units, Contrastive Learning, Modality Gap, PAMAP2
% \end{abstract}

\begin{abstract}

Zero-shot learning (ZSL) for inertial measurement unit (IMU)-based human activity
recognition (HAR) faces a central challenge: bridging the gap between sensor embeddings
and semantic class representations. We systematically evaluate seven configurations
combining three inference methods with two training pipelines on the PAMAP2 dataset,
using 14 seen and 4 unseen activity classes with subjects 108 and 109 held out for
testing. We find that the modality gap is a training-time phenomenon governed by the
encoder objective. A temporal convolutional network (TCN) trained with cross-entropy
over label-name Sentence-BERT prototypes yields sensor embeddings with a mean cosine
similarity of 0.30 to the corresponding text prototypes, while replacing the label-name prototype targets with discriminative 
activity descriptions raises this to 0.69. This alignment improvement transfers consistently across all three inference
methods. The strongest result combines contrastive training with inverted softmax
correction, achieving 73.2\% accuracy and 0.583 macro F1 on unseen classes, compared
to 58.3\% accuracy and 0.34 macro F1 for the label-name baseline. A secondary finding
is that richer text descriptions reduce inter-prototype separability in Sentence-BERT
space, because shared biomechanical vocabulary causes the language model to compress
the prototype cloud. This effect does not negate the benefits of contrastive alignment
provided prototype descriptions retain sufficient discriminative vocabulary. We also
demonstrate that overall accuracy is a misleading primary metric when test-set class
distributions are imbalanced, and recommend macro-averaged F1 as the standard reporting
metric for ZSL-HAR benchmarks.

\vspace{1em}

\end{abstract}

\section{Introduction}
\label{sec:intro}

Human Activity Recognition (HAR) using wearable IMU sensors has applications in healthcare monitoring, assistive technology, and sports science. Supervised HAR systems achieve high accuracy on seen classes but require labeled sensor data for every target activity, a constraint that becomes prohibitive as the set of activities grows or changes. Zero-Shot Learning addresses this by enabling recognition of activities unseen during training, leveraging semantic knowledge transferred through a shared embedding space that connects sensor signals to natural language descriptions of activities.

The central challenge in ZSL-HAR is the \textbf{modality gap}: the sensor embedding space, produced by a temporal encoder trained on IMU windows, and the semantic prototype space, produced by a language model encoding activity descriptions, occupy geometrically separate regions of the shared unit hypersphere. A sensor window labelled ``running'' does not naturally produce an embedding close to the Sentence-BERT encoding of the word ``running'', because the two modalities require explicit geometric alignment to share a meaningful space.

We make the following contributions:

\begin{itemize}
    \item An empirical demonstration that contrastive semantic training, 
    which replaces label-name Sentence-BERT prototypes with discriminative 
    activity descriptions as training targets, reduces the mean text-sensor 
    cosine gap from 0.30 to 0.69 and drives consistent improvements across 
    all three inference methods evaluated.

    \item A counterintuitive finding that richer prototype descriptions 
    reduce inter-prototype separability in Sentence-BERT space, because 
    shared biomechanical vocabulary causes the language model to compress 
    the prototype cloud. Despite this, encoder alignment quality dominates 
    downstream performance: a well-aligned encoder navigates a more 
    compressed prototype space more effectively than a misaligned encoder 
    navigates a well-separated one.

    \item An argument that overall accuracy is a misleading primary metric 
    for ZSL-HAR when test sets are class-imbalanced. Macro-averaged F1 
    assigns equal weight to all unseen classes regardless of support, 
    directly measuring whether the model has genuinely transferred knowledge 
    across the full unseen class set, and we recommend it as the standard 
    reporting metric for ZSL-HAR benchmarks.
\end{itemize}

\section{Related Work}

\subsection{Zero-Shot Learning for Sensor-Based HAR}

Zero-shot learning (ZSL) for IMU-based human activity recognition requires a semantic intermediate representation that bridges the seen-to-unseen transfer problem without labelled sensor examples at test time. Early sensor-domain ZSL work relied on shallow, text-derived class descriptors. Al~Machot et al.~\cite{almachot2020} applied Word2Vec embeddings to non-visual smart-home sensor streams and demonstrated that semantic label similarity can support transfer to unseen activity classes without activity-specific feature annotation; however, single-token embeddings lack the capacity to encode biomechanical context, which limits their discriminability when target classes share common vocabulary. Matsuki et al.~\cite{matsuki2019} provided the most directly comparable evaluation to our own, testing GloVe and fastText word vectors as prototypes for IMU-based ZSL on PAMAP2, OPP, and HASC. A key limitation they identified was strong sensitivity to the choice of unseen classes, which they attributed to cases where word-embedding distance between seen and unseen classes does not reflect the true sensor-signal similarity. Both of these works, and the supervised sensor encoders that preceded them, share a common structural weakness: the sensor encoder is trained against label-name SBERT prototypes 
and is therefore never incentivised to enter the regions of the 
embedding space occupied by richer semantic representations. The result is a persistent \textit{modality gap} between the two independently trained spaces. Our work directly targets this failure mode by replacing the cross-entropy-over-indices objective with a contrastive loss that uses Sentence-BERT~\cite{sbert} class prototypes as training targets, imposing alignment as an explicit training-time objective.

More recent work has shifted to richer auxiliary modalities to supply semantic supervision. Tong et al.~\cite{tong2021} (VbZSL) used I3D video features as class prototypes, reasoning that video encodes motion-specific information that single-word labels cannot convey, and reported substantially improved unseen-class transfer relative to text-only baselines. TEZARNet~\cite{tezarnet} combined I3D video prototypes with a bidirectional LSTM IMU encoder and a temporal alignment mechanism, achieving state-of-the-art results across PAMAP2, DaLiAc, UTD-MHAD, and MHEALTH. SEZ-HARN~\cite{sezharn} extended the TEZARNet architecture with a self-explainability component and matched its accuracy while providing per-class attribution. Despite strong performance, all video-based approaches share a practical constraint: they require curated activity video recordings for every target class, which limits their applicability when new activities must be added without access to recording infrastructure. Our text-only approach addresses this constraint by relying solely on natural language descriptions, at the acknowledged cost of lower prototype separability for activities whose biomechanical signatures cannot be fully distinguished through language.

\subsection{Contrastive Sensor--Language Alignment}

CLIP~\cite{clip} established the principle that jointly training dual encoders with a contrastive objective produces a shared embedding space suitable for zero-shot classification, rendering post-hoc alignment unnecessary. The critical implication for the sensor domain is that geometric alignment should be treated as a training-time objective rather than an inference-time correction. Prior sensor-ZSL work consistently missed this opportunity: even when rich semantic prototypes were available, sensor encoders were trained with standard classification losses and aligned to the prototype space only at test time.

IMU2CLIP~\cite{imu2clip} was the first to apply CLIP-style contrastive pre-training directly to wearable IMU data, projecting sensor recordings into the joint CLIP representation space using paired IMU, video, and text from the large-scale Ego4D dataset. IMU2CLIP validates the alignment-at-training-time principle for the sensor domain, but its reliance on hundreds of thousands of paired multimodal examples places it out of reach for standard HAR benchmarks including PAMAP2. UniMTS~\cite{unimts} (NeurIPS 2024) further generalised contrastive motion pre-training across sensor locations and mounting orientations using LLM-enriched text descriptions, achieving a 340\% zero-shot improvement over prior baselines across 18 benchmark datasets. However, recent work has noted that UniMTS retains limited semantic granularity for complex or fine-grained activity classes~\cite{zara2025}, precisely the setting evaluated in the present study. Neither IMU2CLIP nor UniMTS examined how prototype description quality interacts with encoder alignment when training is confined to a single small dataset, which is the central question of our ablation study.

Haresamudram et al.~\cite{haresamudram2024} (AAAI 2025) investigated the limitations of natural language supervision for wearable HAR in a controlled study and found that naive CLIP-style contrastive pre-training performs substantially worse than standard supervised or self-supervised baselines on sensor data. They identify sensor heterogeneity and insufficient richness of activity text descriptions as the primary causes. Our results reproduce the core finding: plain label-name prototypes yield a mean text-sensor cosine similarity of only 0.30, confirming that single-word supervision is insufficient for encoder alignment. Our work extends their analysis by ablating the specific form of richer description and identifying a non-trivial trade-off that they did not report: richer biomechanical descriptions improve encoder alignment but simultaneously compress prototype separability, raising the mean off-diagonal cosine similarity from 0.29 to 0.50, while discriminative vocabulary-specific descriptions achieve the better balance between both objectives.

Ji et al.~\cite{hargpt} (FMSys 2024) explored a complementary direction by prompting GPT-4 directly with downsampled IMU readings for zero-shot HAR and found that, with appropriate role-play prompting, large language models (LLMs) can match or exceed traditional baselines on datasets with well-separated activity classes. This approach avoids the encoder alignment problem entirely but depends on closed-model API access and degrades when inter-class similarity is high, which is precisely the regime in which our contrastive encoder with discriminative descriptions achieves its largest gains. SensorLLM~\cite{sensorllm} (EMNLP 2025) addresses alignment from a different angle, coupling a pre-trained LLM with wearable sensor channels via a two-stage fine-tuning framework supervised by automatically generated trend-descriptive text. Their finding that human-interpretable, channel-level descriptions are a prerequisite for effective LLM-based sensor reasoning independently corroborates our conclusion that prototype description quality is a first-order variable in any text-grounded sensor recognition system.

\subsection{Understanding the Modality Gap}

Liang et al.~\cite{modalitygap} (NeurIPS 2022) formally characterised the modality gap in multi-modal contrastive models, showing that independently initialised encoders confine their outputs to geometrically separate cones on the unit hypersphere. They attribute this to the cone effect at initialisation, in which the contrastive loss preserves the separation between modality-specific regions rather than eliminating it, and they demonstrate empirically that the magnitude of the gap directly influences downstream zero-shot classification performance. This theoretical framework explains the core problem observed in prior sensor-ZSL work: a TCN encoder trained with label-name SBERT prototypes and a 
Sentence-BERT encoder trained on text occupy separate cones because 
neither objective encourages cross-modal co-location. Our empirical measurement of the text-sensor cosine gap before and after contrastive training, from 0.30 in the baseline to 0.69 in the best configuration, constitutes a direct quantification of the phenomenon Liang et al.\ describe, applied for the first time to the IMU-text modality pair on PAMAP2. Notably, Liang et al.\ also found that some residual gap can benefit zero-shot performance by preserving within-modality structure, which is consistent with our finding that discriminative prototype separability matters alongside alignment quality and that completely collapsing the gap is not the goal.

\subsection{Evaluation Metrics in Zero-Shot HAR}

The choice of evaluation metric is a largely overlooked methodological issue in ZSL-HAR that can substantially affect the conclusions drawn from experimental results. Overall classification accuracy, the metric reported in the majority of 
prior ZSL-HAR studies including Matsuki et al.~\cite{matsuki2019}, 
VbZSL~\cite{tong2021}, and TEZARNet~\cite{tezarnet}, is sensitive to 
class imbalance in the test set. When one unseen class dominates the test distribution, a classifier that learns to recognise only that class will achieve misleadingly high accuracy while failing on all others. Xian et al.~\cite{xian2018} raised a related concern in the visual ZSL literature, demonstrating that inconsistent evaluation protocols, including imbalanced class splits and mixed seen-unseen test sets, rendered published results incomparable across studies and led to systematic overestimation of model performance. In the PAMAP2 test partition used in this study, folding laundry accounts for 51\% of test windows; a model predicting only that class would achieve approximately 50\% overall accuracy while performing at chance on all remaining unseen classes. We therefore adopt macro-averaged F1 as the primary evaluation metric throughout, assigning equal weight to all four unseen classes regardless of support, and demonstrate that overall accuracy and macro F1 lead to qualitatively different rankings across configurations. We recommend that future ZSL-HAR benchmarks on class-imbalanced datasets adopt macro-averaged F1 as the standard reporting metric.

\subsection{Hubness and Inference-Time Corrections in ZSL}

In high-dimensional embedding spaces, the hubness problem causes a small number of prototype points to attract a disproportionately large share of nearest-neighbour assignments regardless of the content of individual queries~\cite{hubness}. This is a structural consequence of distance concentration in high-dimensional spaces and is particularly damaging in ZSL settings where each unseen class is represented by a single prototype point. Smith et al.~\cite{smith2017isf} introduced the Inverted Softmax (ISF) 
as an inference-time correction, normalising each prototype's similarity 
score by its total accumulated score across all test queries so as to 
penalise prototypes that attract queries indiscriminately. In prior sensor-ZSL work, ISF has been applied as a primary performance lever, implicitly treating inference-time correction as a substitute for training-time alignment. Our results challenge this prioritisation directly. In the baseline pipeline, ISF applied over a poorly aligned encoder 
produces marginal improvements in macro F1 and degrades accuracy in 
some configurations. In the contrastive pipeline, the same ISF 
correction applied over a well-aligned encoder produces the largest 
single performance gain reported in this paper, raising macro F1 
from 0.42 to 0.58. The N-occurrence skewness analysis in Section~6.5 further confirms that no configuration exceeds the conventional hubness threshold of skewness greater than 1.0, indicating that the benefit of ISF in our experiments reflects redistribution of assignment pressure caused by residual modality gap rather than correction of classical geometric hubness. The implication is that inference-time corrections and training-time alignment are not substitutes: the effectiveness of corrections such as ISF is gated by the quality of the underlying encoder.

\section{Dataset and Experimental Setup}
\label{sec:dataset}

\subsection{PAMAP2 Dataset and Class Split}
\label{subsec:pamap2}

PAMAP2 \cite{pamap2} contains IMU recordings from nine subjects (subject101 to subject109) performing up to 18 activities at 100~Hz. Each subject wears three IMU units, on the hand, chest, and ankle, that record accelerometer, gyroscope, and magnetometer signals. We retain 31 feature channels after removing the accelerometer channels that saturate during high-intensity activities and the orientation quaternions, which are unreliable across recording sessions.

We use a fixed 14/4 seen/unseen class split that is held constant across all configurations.

\textbf{Seen classes (14):} lying, sitting, standing, walking, cycling, Nordic walking, watching TV, computer work, car driving, ascending stairs, ironing, house cleaning, playing soccer, and rope jumping.

\textbf{Unseen classes (4):} running (ID~5), descending stairs (ID~13), vacuum cleaning (ID~16), and folding laundry (ID~18).

\textbf{Test subjects:} subject108 and subject109 are held out entirely and contribute no data to training or validation. Playing soccer is nominally a seen class, but it has zero training windows under this split and is therefore excluded from alignment training (Approach~3, \cref{subsec:alignment}) through an explicit zero-window guard, leaving 13 valid seen-class training pairs.

\subsection{Windowing and Data Splits}
\label{subsec:windowing}

Training windows span 500 frames (5~seconds at 100~Hz) with 50\% overlap, and test windows span 1{,}000 frames (10~seconds) with 50\% overlap. We apply a purity threshold requiring at least 85\% majority-class frames per window. Feature values are Z-scored using training-set statistics only, with no information from the validation or test partitions used in normalisation.

\begin{table}[t]
\centering
\caption{Dataset split sizes.}
\label{tab:split}
\begin{tabular}{lr}
\toprule
Split & Windows \\
\midrule
Training & 5{,}930 \\
Validation (15\% of training) & 1{,}046 \\
Test (unseen classes, held-out subjects) & 194 \\
\bottomrule
\end{tabular}
\end{table}

\subsection{Class Imbalance and the Case for Macro F1}
\label{subsec:imbalance}

The 194 test windows are severely imbalanced across the four unseen classes, as shown in \cref{tab:test-dist}.

\begin{table}[t]
\centering
\caption{Test-set class distribution.}
\label{tab:test-dist}
\begin{tabular}{lrr}
\toprule
Class & Windows & Proportion \\
\midrule
Running & 31 & 16.0\% \\
Descending stairs & 17 & 8.8\% \\
Vacuum cleaning & 47 & 24.2\% \\
Folding laundry & 99 & 51.0\% \\
\bottomrule
\end{tabular}
\end{table}

Folding laundry alone accounts for 51\% of test windows. A model that correctly classifies only folding laundry while assigning the remaining three classes at random would reach roughly 50\% overall accuracy, far above the 25\% random-chance baseline for four classes. Overall accuracy therefore conflates genuine zero-shot generalisation with performance on a single dominant class.

We adopt macro-averaged F1 as the primary metric throughout, assigning equal weight to all four unseen classes regardless of support. Overall accuracy is reported for comparability with prior work but should not be interpreted in isolation. Balanced accuracy, defined as the unweighted mean of per-class recall, is additionally reported for the contrastively trained configurations that use discriminative descriptions.

\subsection{Semantic Encoder}
\label{subsec:encoder}

All class prototypes are encoded with \texttt{all-mpnet-base-v2} from Sentence-BERT \cite{sbert}, producing 768-dimensional L2-normalised vectors. We evaluate three prototype text formulations across our experiments: label names (e.g., ``running''), rich descriptions (biomechanical sentences), and discriminative descriptions (class-exclusive vocabulary).

\section{Architecture and Training}
\label{sec:arch}

\subsection{TCN Encoder}
\label{subsec:tcn}

A Temporal Convolutional Network maps sensor windows to 768-dimensional L2-normalised embeddings. The encoder consists of three dilated residual TCN blocks with channel widths of 64, 96, and 128, dilation rates of 1, 2, and 4, and kernel size 5, followed by global average pooling and a two-layer projection head mapping from 128 to 512 to 768 dimensions. L2 normalisation is applied to the final output. The total parameter count is approximately 0.73M. The architecture is identical across all experiments; differences between configurations arise solely from the training objective and the prototype text formulation.

\subsection{Training Objectives}
\label{subsec:objectives}

\textbf{Baseline encoder: label-name prototypes.} The baseline encoder, used by Approaches~1 to~3, is trained with cross-entropy over semantic logits using label-name SBERT prototypes as targets, with temperature $\tau = 0.07$. The loss is identical in form to contrastive semantic training, but the prototype targets carry minimal semantic structure because they are derived from single-word activity labels. The training objective is given in \cref{eq:contrastive_loss}.

\begin{equation}
    \mathcal{L} = -\frac{1}{N}\sum_{i=1}^{N} \log 
    \frac{\exp(\mathbf{e}_i^\top \mathbf{s}_{y_i}/\tau)}
    {\sum_{c=1}^{C} \exp(\mathbf{e}_i^\top \mathbf{s}_c/\tau)}
    \label{eq:contrastive_loss}
\end{equation}

\textbf{Contrastive semantic training.} The same loss is applied with richer SBERT prototype targets: rich biomechanical descriptions for the encoder used by Approaches~4a and~4b, and discriminative descriptions for the encoder used by the modified approaches (A1-M, A2-M, and A3-M). This objective directly maximises the agreement between each sensor embedding and its corresponding class-specific semantic prototype across all seen classes simultaneously. It is the IMU analogue of CLIP-style contrastive alignment applied to the sensor domain.

Training hyperparameters differ across encoder configurations. The baseline encoder uses AdamW with learning rate $2\times10^{-3}$, weight decay $1\times10^{-4}$, cosine annealing over 15 epochs, and early stopping with patience 10. The contrastive encoder trained with rich descriptions reduces the learning rate to $1\times10^{-3}$, increases the maximum number of epochs to 40, and increases patience to 15. The contrastive encoder trained with discriminative descriptions uses the same configuration as the rich-description encoder: AdamW with learning rate $1\times10^{-3}$, weight decay $1\times10^{-4}$, cosine annealing over 40 epochs, and early stopping with patience 15. Gradient norms are clipped to 1.0 in all configurations, and the best checkpoint on validation nearest-prototype accuracy is restored for inference.

\begin{figure}[t]
    \centering
    \includegraphics[width=\linewidth]{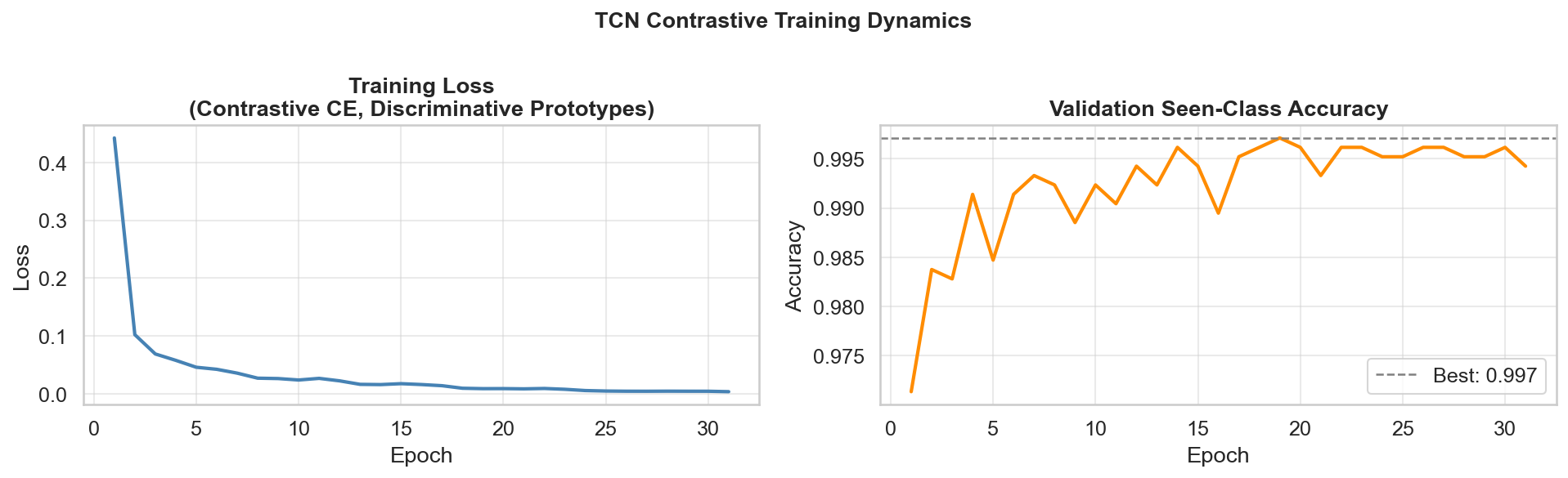}
    \caption{Training dynamics for the contrastive encoder with discriminative descriptions: training loss (left) and validation seen-class accuracy (right).}
    \label{fig:training_curve}
\end{figure}

\section{ZSL Inference Approaches}
\label{sec:inference}

The three inference approaches described below are evaluated under two encoder configurations. The implementations are identical in method but differ in two important ways. First, the encoder producing the sensor embeddings differs: the baseline approaches (A1 to A3) use a TCN trained with label-name prototypes, while the contrastive approaches (A4a, A4b, and the modified approaches A1-M to A3-M) use a contrastively trained TCN with richer prototype targets. Second, the ISF temperature sweep is optimised on overall accuracy for the baseline approaches and on macro F1 for the modified approaches, reflecting the shift to macro F1 as the primary metric in the modified pipeline. Note that for A2 and A3, the 
baseline encoder trained on label-name prototypes is evaluated at 
inference against rich-description prototypes; this isolates the 
contribution of inference-time corrections from training-time alignment 
and explains the prototype column entries in \cref{tab:main_results}

\subsection{Approach 1: Nearest-Prototype Cosine Matching}
\label{subsec:cosine}

The simplest ZSL inference strategy assigns each test embedding to the unseen class whose SBERT prototype has the highest cosine similarity:

\begin{equation}
    \hat{y} = \arg\max_{c \in \mathcal{U}} \mathbf{e}^\top \mathbf{s}_c
    \label{eq:cosine_nn}
\end{equation}

In the baseline configuration (A1) this uses label-name prototypes against baseline TCN embeddings; in the modified configuration (A1-M) it uses discriminative-description prototypes against contrastive TCN embeddings. Because both the encoder and the prototype type change between the two, this single inference method serves as a controlled comparison of everything that changed in the training pipeline.

\subsection{Approach 2: ZSL with ISF and Cross-Modal Centering}
\label{subsec:isf}

Two corrections are applied on top of the embedding space. First, cross-modal centering computes a global mean over all $N$ test sensor embeddings and all 18 class prototypes (both seen and unseen) combined, then subtracts this mean from both modalities and renormalises onto the unit sphere:

\begin{equation}
    \tilde{\mathbf{e}}_i = \frac{\mathbf{e}_i - \bar{\mu}}{\|\mathbf{e}_i - \bar{\mu}\|}, \quad
    \tilde{\mathbf{s}}_c = \frac{\mathbf{s}_c - \bar{\mu}}{\|\mathbf{s}_c - \bar{\mu}\|}
    \label{eq:centering}
\end{equation}
\begin{equation}
    \bar{\mu} = \frac{1}{N+C}\left(\sum_{i=1}^{N}\mathbf{e}_i + 
    \sum_{c=1}^{C}\mathbf{s}_c\right)
    \label{eq:global_mean}
\end{equation}

with $N$ test sensor embeddings and $C = 18$ total class prototypes.

Second, Inverted Softmax (ISF) normalises each prototype's score by its total accumulated score across all test queries, penalising prototypes that attract many queries regardless of content, as given in \cref{eq:isf}:

\begin{equation}
    \hat{y} = \arg\max_{c \in \mathcal{U}} 
    \frac{\exp(\tilde{\mathbf{e}}^\top \tilde{\mathbf{s}}_c / \beta)}
    {\sum_{q=1}^{N} \exp(\tilde{\mathbf{e}}_q^\top \tilde{\mathbf{s}}_c / \beta)}
    \label{eq:isf}
\end{equation}

where $\beta$ is the ISF temperature and the denominator sums over all $N$ test queries for each prototype. The temperature $\beta$ is selected by sweeping over $\{0.05, 0.10, 0.20, 0.30, 0.50, 0.75, 1.00, 1.50, 2.00\}$. For the baseline approach (A2) the sweep is optimised on overall accuracy; for the modified approach (A2-M) it is optimised on macro F1, consistent with the shift to balanced evaluation in the modified pipeline. The modified approach also uses discriminative-description prototypes encoded by the contrastive TCN rather than the label-name prototypes and baseline TCN of A2; the centering and ISF corrections are otherwise identical.

\subsection{Approach 3: Cross-Modal Alignment via Seen-Class Semantic Mapping}
\label{subsec:alignment}

Cross-modal alignment learns a linear map from text prototype space into sensor space using seen-class pairs as supervision. For each seen class with training windows, the mean sensor embedding centroid is computed and paired with its SBERT text prototype. Playing soccer is excluded because it has zero training windows under the held-out subject configuration, yielding 13 valid training pairs.

Both the text prototype matrix $T$ and the sensor centroid matrix $C$ are first compressed into separate $K$-dimensional PCA subspaces fitted independently on each modality. A ridge-regularised alignment matrix $W$ is then fitted between these two independent subspaces, as given in \cref{eq:ridge}:

\begin{equation}
    W^* = (T_{\text{pca}}^\top T_{\text{pca}} + \lambda I)^{-1} 
    T_{\text{pca}}^\top C_{\text{pca}}
    \label{eq:ridge}
\end{equation}

where $T_{\text{pca}}$ and $C_{\text{pca}}$ are the $K$-dimensional PCA-compressed representations of the text prototypes and sensor centroids respectively, with separate PCA bases fitted on each modality independently. This PCA compression avoids the underdetermined regime that would arise from fitting a 768-dimensional map on only 13 data points.

The hyperparameter $K \in \{3, 5, 7, 9, 11\}$ and the ridge coefficient $\lambda$ are selected via leave-one-out cosine reconstruction error on seen classes. The $\lambda$ grids differ slightly between configurations: the baseline approach (A3) uses $\{10^{-3}, 10^{-2}, 0.1, 1.0, 5.0, 10.0, 50.0\}$ and the modified approach (A3-M) uses $\{10^{-3}, 10^{-2}, 0.1, 1.0, 5.0, 10.0\}$.

At inference time each unseen-class text prototype is passed through the alignment pipeline: PCA compression using the text basis, linear transformation through $W$, and decompression using the sensor basis. The projected prototype is L2-renormalised and used for nearest-prototype matching entirely within sensor space. ISF is applied on top with an independently swept temperature, optimised on overall accuracy for the baseline approach (A3) and on macro F1 for the modified approach (A3-M).

\section{Results}
\label{sec:results}

\subsection{Modality Gap Measurement}
\label{subsec:gap-measurement}

The mean cosine similarity between each seen-class sensor centroid and its corresponding SBERT text prototype provides a direct measure of the modality gap. \cref{tab:modality_gap} reports this metric across the three encoder configurations evaluated.

\begin{table}[h]
    \centering
    \caption{Modality gap across encoder configurations. $^\dagger$All cosines are measured against rich-description prototypes as a common reference, regardless of the prototype type used in training.}
    \label{tab:modality_gap}
    \begin{tabularx}{\textwidth}{Xccc}
        \toprule
        \textbf{Encoder Configuration} & \textbf{Mean Text-Sensor Cosine} 
        & \textbf{Min} & \textbf{Max} \\
        \midrule
        Baseline TCN, label-name prototypes & 0.3027 & 0.079 (lying) 
        & 0.510 (Nordic walking) \\
        Contrastive TCN, rich descriptions & 0.6326 & n/a & n/a \\
        Contrastive TCN, discriminative descriptions & 
        \textbf{0.6851} & n/a & n/a \\
        \bottomrule
    \end{tabularx}
\end{table}

\begin{figure}[H]
    \centering
    \includegraphics[width=\linewidth]{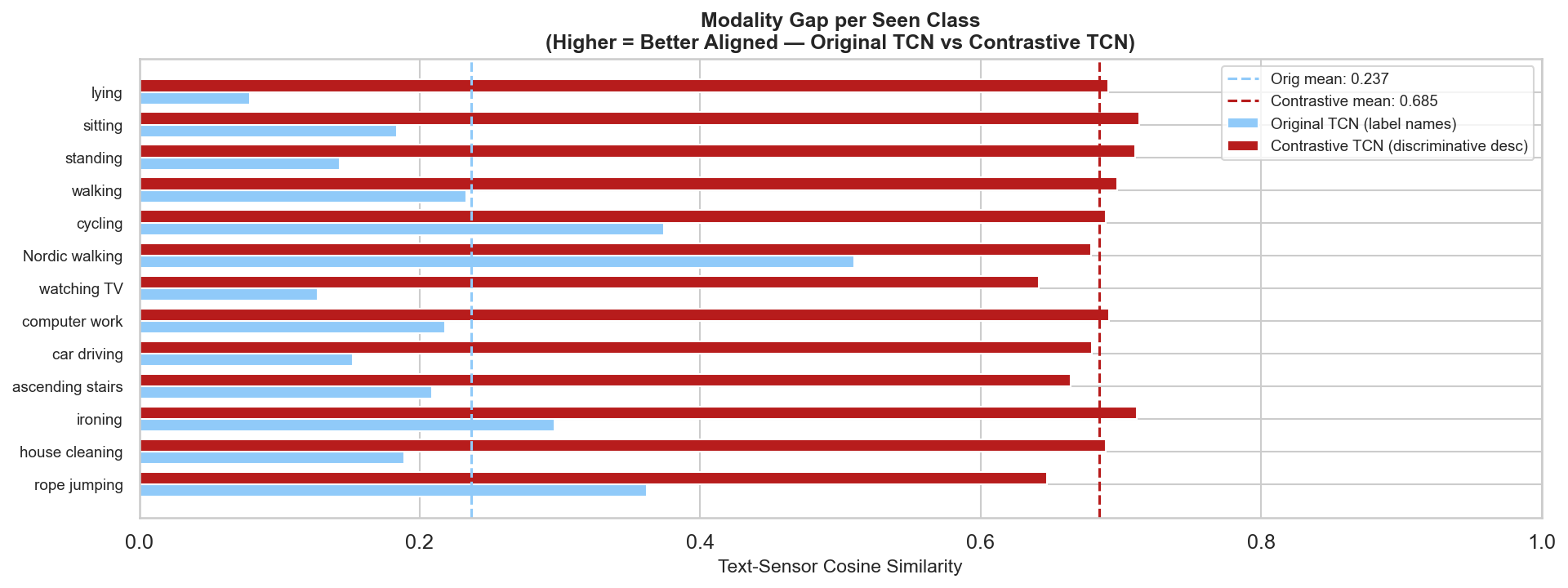}
    \caption{Modality gap per seen class.}
    \label{fig:modality_gap}
\end{figure}

Contrastive training raised the mean text-sensor cosine from 0.3027 in the baseline to 0.6326 with rich biomechanical descriptions, and further to 0.6851 with discriminative descriptions. This confirms that the training objective directly controls the magnitude of the modality gap, and that the choice of prototype description influences the degree of alignment achieved.

\subsection{Prototype Separability}
\label{subsec:separability}

A counterintuitive finding emerged from our prototype engineering efforts. Despite careful vocabulary curation designed to maximise inter-class SBERT distance, both rich and discriminative descriptions made prototype separability worse than plain label names, as shown in \cref{tab:prototype_separability}.

\begin{table}[h]
    \centering
    \caption{Unseen-class prototype separability (mean off-diagonal cosine similarity; lower is more separable).}
    \label{tab:prototype_separability}
    \begin{tabularx}{\textwidth}{Xc}
        \toprule
        \textbf{Prototype Type} & \textbf{Mean Off-Diagonal Cosine} \\
        \midrule
        Label names & \textbf{0.288} \\
        Rich biomechanical descriptions & 0.500 \\
        Discriminative descriptions & 0.438 \\
        \bottomrule
    \end{tabularx}
\end{table}

\begin{figure}[H]
    \centering
    \includegraphics[width=\linewidth]{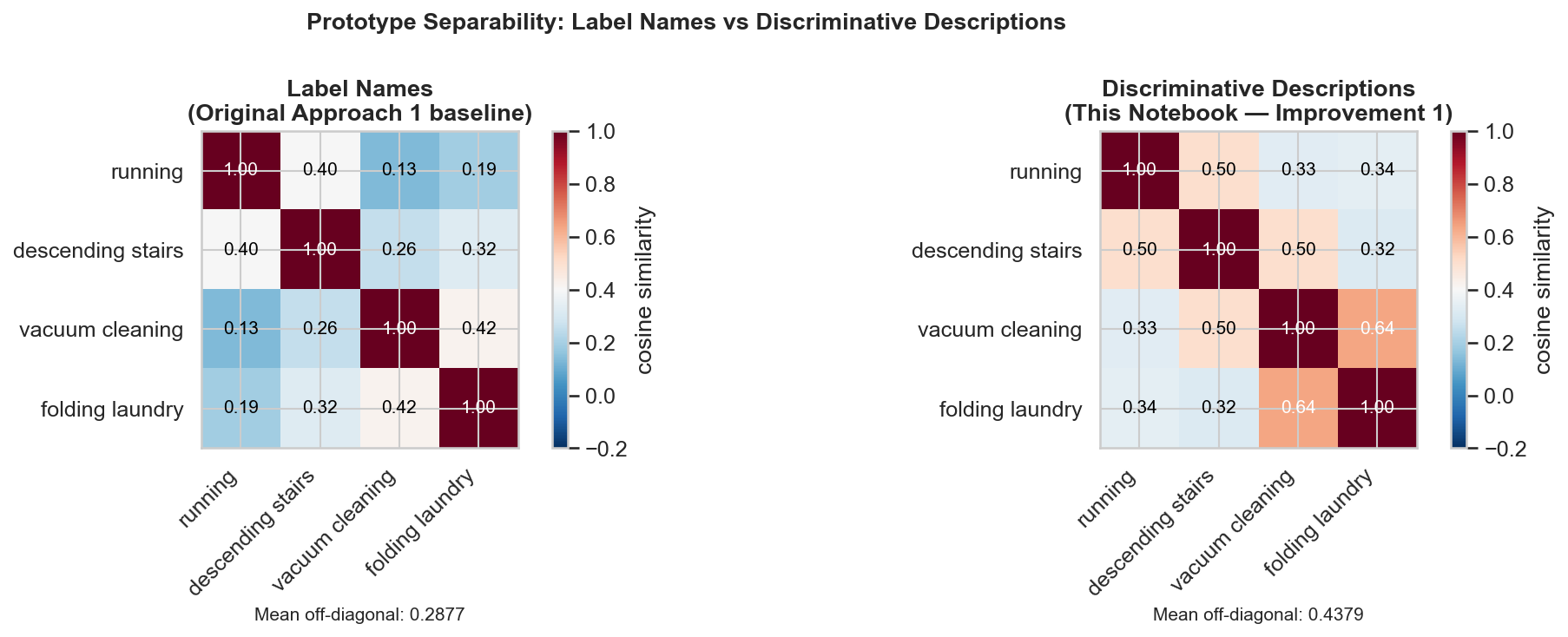}
    \caption{Prototype separability using class labels versus activity descriptions.}
    \label{fig:proto_sep}
\end{figure}

Rich and discriminative descriptions share biomechanical vocabulary, including terms related to acceleration, sensor axes, and body locations, which SBERT interprets as domain-level similarity and uses to compress the prototype cloud. Plain label names share no vocabulary across classes and are therefore more separable in SBERT space. Despite this, the configurations that pair discriminative descriptions with contrastive training (A1-M to A3-M) substantially outperform the label-name configurations across all metrics. This demonstrates that encoder alignment quality dominates prototype geometry: even a more compressed prototype space is navigated effectively when sensor embeddings are genuinely aligned with it.

\subsection{Main Results}
\label{subsec:main-results}

\begin{table}[H]
    \centering
    \caption{ZSL results across all configurations on PAMAP2 (4 unseen classes, 194 test windows). Discrim.\ desc.\ = discriminative descriptions; Bal.\ Acc = balanced accuracy.}
    \label{tab:main_results}
    \resizebox{\textwidth}{!}{%
    \begin{tabular}{clllccc}
        \toprule
        \textbf{Config} & \textbf{Encoder} & 
        \textbf{Prototypes} & \textbf{Inference} & \textbf{Overall Acc} & 
        \textbf{Macro F1} & \textbf{Bal. Acc} \\
        \midrule
        A1 & Baseline & Label names & Cosine NN & 0.58 & 0.34 & n/a \\
        A2 & Baseline & Rich desc. & ISF + centering & 0.55 & 0.35 & n/a \\
        A3 & Baseline & Rich desc. & Alignment + ISF & 0.42 & 0.32 & n/a \\
        A4a & Contrastive & Rich desc. & Cosine NN & 0.40 & 0.29 & n/a \\
        A4b & Contrastive & Rich desc. & ISF + centering & 0.45 & 0.33 & n/a \\
        \midrule
        A1-M & Contrastive & Discrim. desc. & Cosine NN & 0.64 & 0.42 & 0.55 \\
        A2-M & Contrastive & Discrim. desc. & ISF + centering & 
        \textbf{0.73} & \textbf{0.5832} & \textbf{0.6350} \\
        A3-M & Contrastive & Discrim. desc. & Alignment + ISF & 0.72 & 0.5695 & 0.5855 \\
        \midrule
        Random chance & n/a & n/a & n/a & 0.25 & 0.25 & 0.25 \\
        \bottomrule
    \end{tabular}}
\end{table}

\subsection{Per-Class Analysis}
\label{subsec:per-class}

\cref{tab:per_class_recall} provides per-class recall for all configurations, revealing patterns that are obscured by aggregate metrics.

\begin{table}[H]
    \centering
    \caption{Per-class recall across all configurations.}
    \label{tab:per_class_recall}
    \begin{tabularx}{\textwidth}{Xcccc}
        \toprule
        \textbf{Configuration} & \textbf{Running} & 
        \textbf{Desc. Stairs} & \textbf{Vacuum Cleaning} & 
        \textbf{Folding Laundry} \\
        \midrule
        \multicolumn{5}{l}{\textit{Baseline pipeline}} \\
        A1 (baseline) & 0.00 & 1.00 & 0.04 & 0.95 \\
        A2 (ISF) & 0.00 & 1.00 & 0.09 & 0.87 \\
        A3 (alignment) & 0.23 & 0.53 & 0.04 & 0.64 \\
        A4a (contrastive + nearest prototype) & 0.13 & 0.71 & 0.00 & 0.76 \\
        A4b (contrastive + ISF) & 0.16 & 0.71 & 0.02 & 0.71 \\
        \midrule
        \multicolumn{5}{l}{\textit{Modified pipeline}} \\
        A1-M (cosine) & 0.03 & 1.00 & 0.17 & 1.00 \\
        \textbf{A2-M (ISF)} & \textbf{0.74} & \textbf{0.59} & \textbf{0.21} & \textbf{1.00} \\
        A3-M (alignment) & 0.39 & 0.53 & 0.43 & 1.00 \\
        \bottomrule
    \end{tabularx}
\end{table}

% baseline pipeline comparison
\begin{figure}[H]
    \centering
    \includegraphics[width=\linewidth]{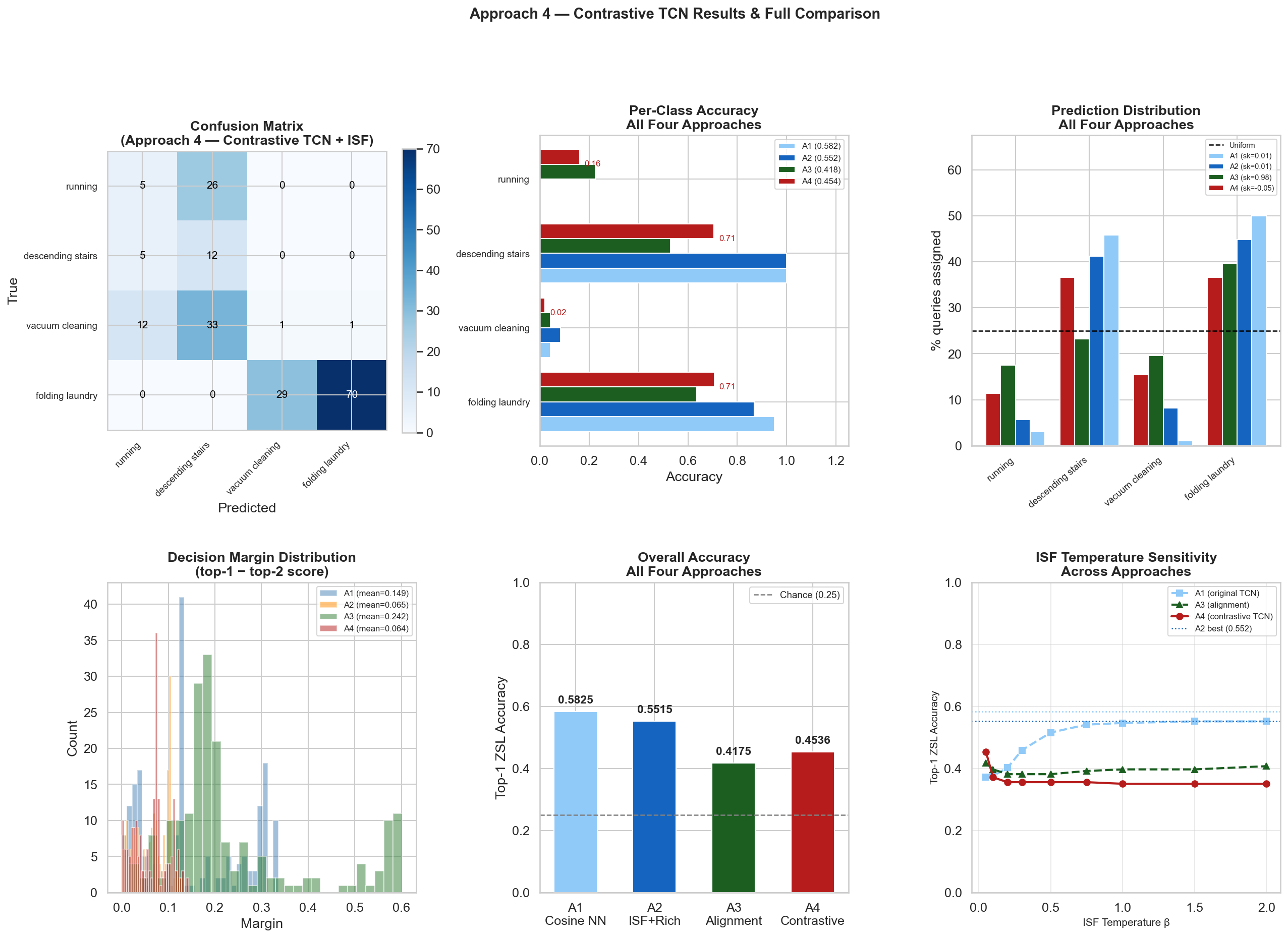}
    \caption{Comparison of all approaches under the baseline pipeline (A1 to A4b).}
    \label{fig:nb1_comparison}
\end{figure}

% modified pipeline comparison
\begin{figure}[H]
    \centering
    \includegraphics[width=\linewidth]{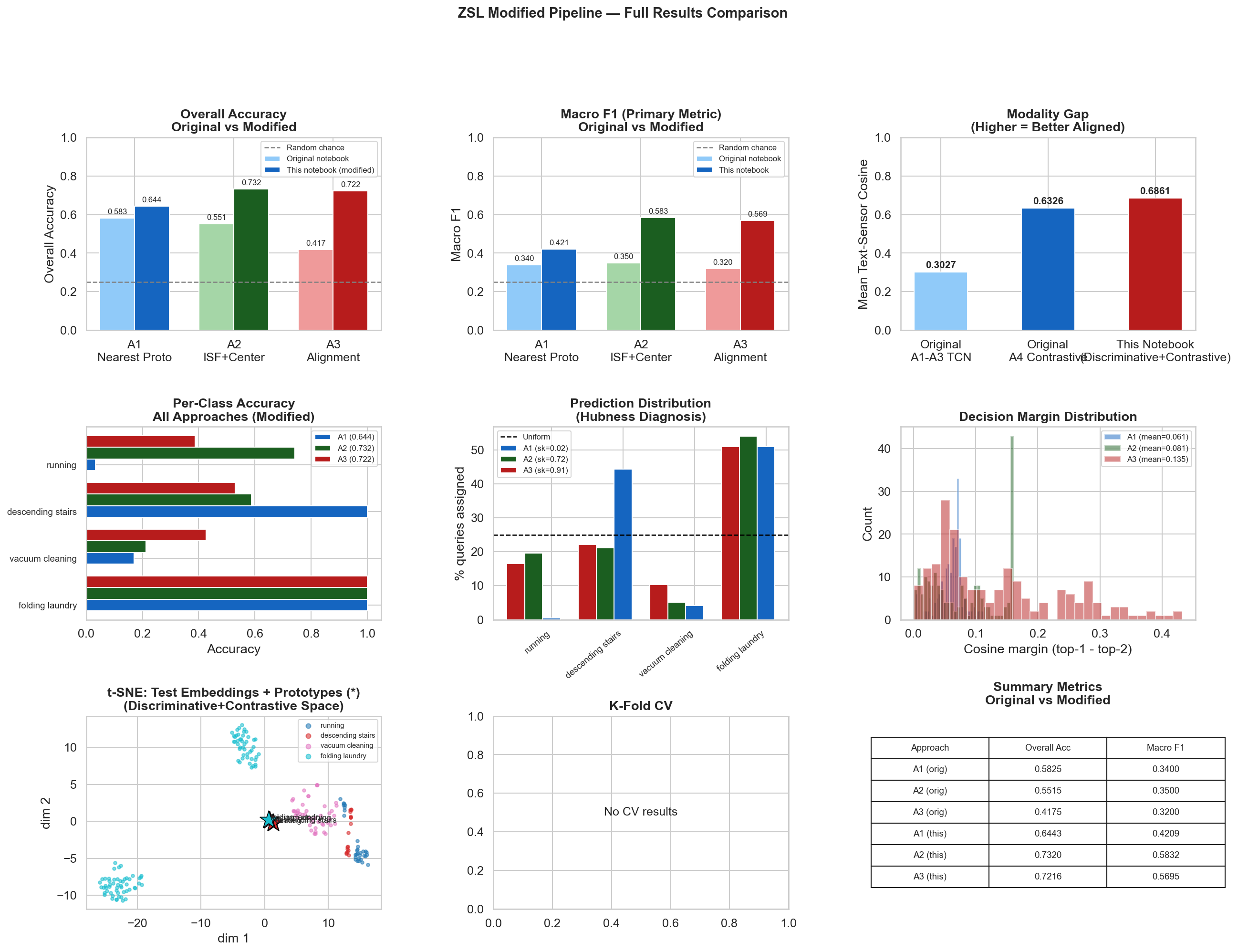}
    \caption{Comparison of all approaches under the modified pipeline (A1-M to A3-M).}
    \label{fig:nb2_comparison}
\end{figure}

Recall on running never exceeds 23\% across the five baseline-pipeline configurations A1 to A4b, reaching 0\% under both nearest-prototype and ISF matching and peaking at 23\% under cross-modal alignment.

Vacuum cleaning remains the hardest class across all configurations, with a maximum recall of 43\% achieved by A3-M. The activity combines slow walking with bilateral arm manipulation, producing sensor signatures that overlap with multiple seen classes. No seen class in the training set has a biomechanically similar profile, which limits the ability of any prototype alignment approach to place its embedding accurately.

\subsection{Prototype Assignment Analysis}
\label{subsec:hubness}

N-occurrence skewness quantifies hubness by measuring the skewness of the distribution of test query assignments across prototypes. \cref{tab:skewness} summarises this across all configurations.

\begin{table}[H]
    \centering
    \caption{N-occurrence skewness (higher values indicate more uneven prototype assignment).}
    \label{tab:skewness}
    \begin{tabularx}{\textwidth}{Xcccccccc}
        \toprule
        \textbf{Configuration} & \textbf{A1} & \textbf{A2} & \textbf{A3} 
        & \textbf{A4a} & \textbf{A4b} & \textbf{A1-M} & \textbf{A2-M} 
        & \textbf{A3-M} \\
        \midrule
        \textbf{Skewness} & 0.009 & 0.007 & 0.976 & NA & $-0.046$ & 0.023 
        & 0.69 & 0.98 \\
        \textbf{Hub class} & none & none & none & NA & none & none & none 
        & none \\
        \bottomrule
    \end{tabularx}
\end{table}

No configuration exceeded the conventional hubness threshold of skewness greater than 1.0, and the baseline cosine approach (A1) explicitly returned no significant hubness. Skewness values for Approaches~1 and~2 are near zero under both encoder configurations (A1, A2, A1-M, A2-M), indicating balanced prototype assignment. Approach~3 under both encoders (A3 and A3-M) shows elevated skewness approaching but not exceeding 1.0, which reflects the natural class imbalance in the test set rather than geometric hubness: folding laundry constitutes 51\% of test windows, so a well-calibrated model is expected to assign a proportionally larger share of queries to it. ISF was applied as a precautionary correction against hubness in the high-dimensional embedding space, and the improvement from A1-M to A2-M (macro F1 from 0.42 to 0.58) is better attributed to redistribution of assignment pressure away from prototypes absorbing queries due to the modality gap, rather than to correction of confirmed hubness.

\begin{figure}[H]
    \centering
    \includegraphics[width=\linewidth]{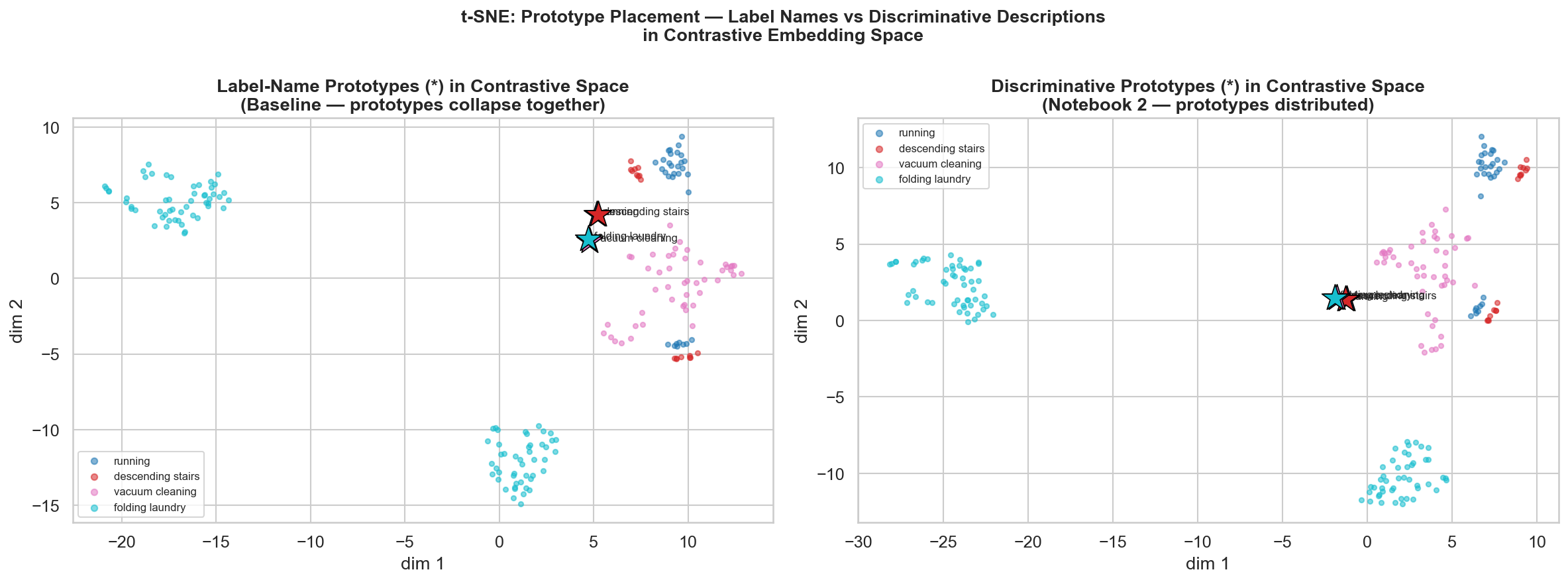}
    \caption{t-SNE prototype placements for class labels versus class descriptions.}
    \label{fig:tsne}
\end{figure}

\subsection{Embedding Space Visualisation}
\label{subsec:tsne}

t-SNE projections of test embeddings with unseen-class prototypes illustrate the effect of contrastive training on prototype placement. In the baseline TCN space, all four unseen-class prototypes collapse into a single cluster far from every sensor embedding cloud, visible in \cref{fig:tsne}. Under the contrastive encoder with discriminative descriptions, the prototype placement becomes more distributed, with the running and descending stairs prototypes moving substantially closer to their corresponding sensor clouds compared with the baseline. Vacuum cleaning and folding laundry prototypes show partial but incomplete displacement toward their clusters, consistent with their lower recall values in \cref{tab:per_class_recall}.

% prototype centroid similarity
\begin{figure}[H]
    \centering
    \includegraphics[width=\linewidth]{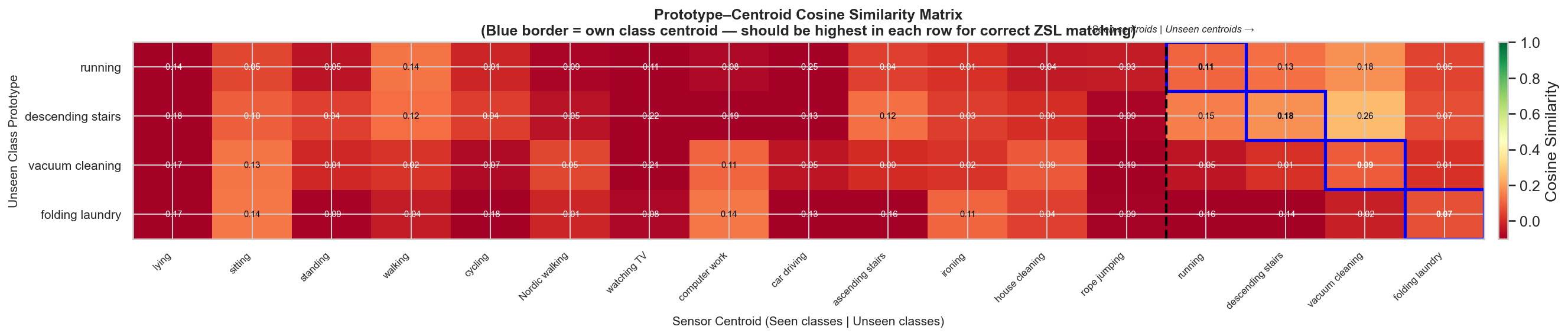}
    \caption{Prototype centroid similarity.}
    \label{fig:proto_centroid}
\end{figure}

In a well-aligned space the diagonal entry for each unseen-class 
row should be the highest in that row. Running and descending stairs 
satisfy this under the contrastive encoder, consistent with their 
stronger recall in A2-M. Vacuum cleaning and folding laundry show 
weaker diagonal dominance, consistent with their lower and 
ceiling-bounded recall in Table~\ref{tab:per_class_recall}.

% tsne

\section{Discussion}
\label{sec:discussion}

\subsection{The Training Objective Is the Dominant Variable}
\label{subsec:dominant-variable}

The most important result of this study is visible in \cref{tab:main_results}. The contrastive encoder paired with the simplest possible inference method (cosine nearest-prototype, A1-M: 64.4\% accuracy, 0.42 macro F1) outperforms the baseline encoder paired with the most sophisticated inference method tested (cross-modal alignment with ISF, A3: 41.8\% accuracy, 0.32 macro F1). This confirms that no amount of inference-time correction can compensate for training-time misalignment. Conversely, a well-aligned encoder substantially reduces the burden on the inference stage: even simple cosine matching under the contrastive encoder (A1-M) outperforms all five baseline-pipeline configurations (A1 to A4b).

\subsection{Why Contrastive Training Alone Was Insufficient with Rich Descriptions}
\label{subsec:why-contrastive-insufficient}

An apparent contradiction in the results requires explanation. Approach~4b, the best-performing contrastive variant under the baseline pipeline, used contrastive training with rich descriptions (mean cosine 0.6326) but achieved only 0.33 macro F1, marginally below the label-name baseline (A1) at 0.34. This occurred because contrastive training in that configuration used rich biomechanical descriptions as prototype targets. These descriptions degraded prototype separability, raising the off-diagonal cosine from 0.288 to 0.500, while simultaneously aligning the encoder to that compressed semantic space. The encoder therefore learned to align with a poorly separated prototype distribution, reducing its ability to discriminate between similar unseen classes. This interaction between encoder alignment quality and prototype separability explains why adding contrastive training on top of poorly separated prototypes does not yield improvements. Using discriminative descriptions with contrastive training (A1-M to A3-M) produced sufficient residual separability to overcome this interaction, yielding consistent improvements across all inference methods despite the descriptions still being less separable than plain label names.

\subsection{Limitations}
\label{subsec:limitations}

\textbf{Single subject split.} All results derive from a single held-out subject pair (subject108 and subject109), yielding 194 test windows. A 4-fold subject-level cross-validation was attempted but could not complete because not all subjects recorded all four unseen activities, a fundamental constraint of PAMAP2. Results should therefore be interpreted as single-split estimates rather than stable cross-validated means.

\textbf{Four unseen classes.} With only four unseen classes, random chance is 25\%, and the imbalance caused by folding laundry at 51\% of test windows substantially inflates overall accuracy, which is why macro F1 is the primary metric throughout.

\textbf{Computational constraints.} All experiments were conducted under limited computational resources, which restricted the scope of the hyperparameter search, the number of training seeds evaluated, and the range of encoder architectures explored. In particular, the TCN encoder architecture and the training hyperparameters reported in \cref{sec:arch} were selected from a constrained sweep rather than an exhaustive grid search, and results are reported for a single training run per configuration rather than as means over multiple random seeds. These constraints preclude strong claims about the stability of individual performance figures. However, the central findings of this study concern \textit{relative} differences between configurations that are large in magnitude (for example, a macro F1 improvement from 0.34 to 0.58 when switching from the baseline to contrastive training with ISF), which are unlikely to be reversed by variance across seeds. The conclusions regarding the training objective as the dominant variable, and the interaction between prototype description quality and encoder alignment, rest on consistent directional patterns observed across all seven configurations rather than on the precise value of any single metric.

\textbf{Text-only semantic prototypes.} The best-performing published ZSL-HAR methods, TEZARNet and SEZ-HARN, use I3D video features as class prototypes, which carry substantially richer motion information than text descriptions. Our approach is complementary in that it requires no activity video data, but it cannot claim parity with video-based approaches on a class-for-class basis.

\textbf{Vacuum cleaning recall ceiling.} No configuration achieves more than 43\% recall on vacuum cleaning. This activity blends locomotion and arm-manipulation signals with no close seen-class analogue, which fundamentally limits the transferability of seen-class knowledge to it.

\subsection{Future Directions}
\label{subsec:future}

Two directions follow most directly from the limitations established above. The most impactful is k-fold class-split evaluation following the TEZARNet and SEZ-HARN protocol, in which the activity super-class held out as unseen is rotated across folds. This would replace the single-split estimates reported here with stable means and confidence intervals, and would enable direct numerical comparison against published baselines on the same partitions. On the modelling side, the unilateral alignment demonstrated here, in which only the sensor encoder is trained against fixed text prototypes, could be extended to joint contrastive training of the TCN and a trainable text encoder, allowing both modalities to move toward a shared space rather than forcing the sensor encoder to chase a frozen and poorly separated prototype cloud. This bilateral objective directly targets the separability collapse documented in \cref{subsec:separability}, since the text encoder would no longer be constrained to the off-the-shelf SBERT geometry. A complementary generative route is to synthesise unseen-class sensor embedding distributions, for example with an aligned variational autoencoder \cite{cadavae}, converting the zero-shot problem into standard supervised classification and removing the single point-estimate prototype assumption of Approaches~1 to~3. We note, however, that generative methods condition on the same semantic prototypes and therefore inherit any prototype-placement error, so this route is unlikely to lift activities such as vacuum cleaning that lack a biomechanically analogous seen class.

The harder problems, the vacuum cleaning recall ceiling and the prototype separability collapse, both stem from the limited expressiveness of static SBERT text prototypes, and recent work points to richer semantic supervision as the most promising remedy. LLM-enriched activity descriptions, as used by UniMTS \cite{unimts}, and LLM-guided semantic alignment approaches \cite{su2025} generate more discriminative class representations than hand-written sentences, which may relieve the vocabulary-sharing effect we observed without sacrificing alignment. A separate line of work treats a large language model as the recogniser itself, reasoning directly over sensor signals \cite{hargpt} or over retrieved evidence \cite{zara2025}, which sidesteps the fixed-prototype assumption entirely. These directions are promising rather than settled: LLM-based recognisers can degrade below random chance on fine-grained activities with highly similar inter-class signatures \cite{xu2025}, which is precisely the regime in which vacuum cleaning is difficult, so their benefit for the hardest classes remains an open empirical question well suited to the PAMAP2 split studied here.

\section{Conclusion}

We have presented a systematic investigation of seven ZSL configurations 
for IMU-based human activity recognition on PAMAP2, covering three 
inference methods and two encoder configurations. Our central finding is
that the modality gap between sensor and semantic embedding spaces is a
training-time phenomenon driven by the encoder objective function, and
that contrastive semantic training using Sentence-BERT class prototypes
as targets raises the mean text-sensor cosine similarity from 0.30 to
0.69, enabling consistent improvements across all inference methods.

The best configuration, which combines contrastive training with
Inverted Softmax correction, achieves 73.2\% overall accuracy and 0.583
macro F1 on four unseen activity classes, compared to 58.3\% accuracy
and 0.34 macro F1 in the label-name baseline. Running recall improves
from 0\% in the baseline nearest-prototype configuration to 74\% in the
best modified configuration, confirming genuine ZSL generalisation. We
also demonstrate that overall accuracy is a misleading primary metric
when test sets are class-imbalanced, and we argue that macro-averaged F1
should be adopted as the standard reporting metric for ZSL-HAR
benchmarks.

A secondary counterintuitive finding is that richer prototype
descriptions consistently reduce SBERT prototype separability relative
to plain label names. This suggests that encoder alignment quality is
the dominant variable determining ZSL performance in the IMU setting,
and that prototype description quality plays a secondary role that can
be actively harmful when the encoder is not well aligned.

\bibliographystyle{plain}
\bibliography{reference}

\begin{thebibliography}{10}

\bibitem{almachot2020}
Fadi Al~Machot, Mohammed~R. Elkobaisi, and Kyandoghere Kyamakya.
\newblock Zero-shot human activity recognition using non-visual sensors.
\newblock {\em Sensors}, 20(3):825, 2020.

\bibitem{sezharn}
D.Y. De~Silva et~al.
\newblock {SEZ-HARN}: Self-explainable zero-shot human activity recognition network.
\newblock {\em arXiv preprint arXiv:2507.00050}, 2025.

\bibitem{tezarnet}
P.N. Deelaka et~al.
\newblock {TEZARNet}: {TE}mporal zero-shot activity recognition network.
\newblock In {\em Neural Information Processing, ICONIP 2023}, volume 1969 of {\em Communications in Computer and Information Science}, Singapore, 2024. Springer.

\bibitem{haresamudram2024}
Harish Haresamudram, Apoorva Beedu, Mashfiqui Rabbi, Sougata Saha, Irfan Essa, and Thomas Ploetz.
\newblock Limitations in employing natural language supervision for sensor-based human activity recognition---{And} ways to overcome them.
\newblock In {\em Proceedings of the AAAI Conference on Artificial Intelligence}, 2025.
\newblock arXiv:2408.12023.

\bibitem{hargpt}
Sijie Ji, Xinzhe Zheng, and Chenshu Wu.
\newblock {HARGPT}: Are {LLM}s zero-shot human activity recognizers?
\newblock In {\em 2024 IEEE International Workshop on Foundation Models for Cyber-Physical Systems \& Internet of Things (FMSys)}, pages 38--43. IEEE, 2024.

\bibitem{zara2025}
Zechen Li, Baiyu Chen, Hao Xue, and Flora~D. Salim.
\newblock {ZARA}: Training-free motion time-series reasoning via evidence-grounded {LLM} agents.
\newblock {\em arXiv preprint arXiv:2508.04038}, 2025.

\bibitem{sensorllm}
Zechen Li et~al.
\newblock {SensorLLM}: Aligning large language models with motion sensors for human activity recognition.
\newblock {\em arXiv preprint arXiv:2410.10624}, 2024.
\newblock To appear at EMNLP 2025.

\bibitem{modalitygap}
Weixin Liang, Yuhui Zhang, Yongchan Kwon, Serena Yeung, and James~Y. Zou.
\newblock Mind the gap: Understanding the modality gap in multi-modal contrastive representation learning.
\newblock In {\em Advances in Neural Information Processing Systems}, volume~35, 2022.

\bibitem{matsuki2019}
Moe Matsuki, Paula Lago, and Sozo Inoue.
\newblock Characterizing word embeddings for zero-shot sensor-based human activity recognition.
\newblock {\em Sensors}, 19(22):5043, 2019.

\bibitem{imu2clip}
Seungwhan Moon, Andrea Madotto, Zhaojiang Lin, Aparajita Saraf, Amy Bearman, and Babak Damavandi.
\newblock {IMU2CLIP}: Language-grounded motion sensor translation with multimodal contrastive learning.
\newblock In {\em Findings of the Association for Computational Linguistics: EMNLP 2023}, pages 13246--13253, Singapore, 2023. Association for Computational Linguistics.

\bibitem{clip}
A.~Radford et~al.
\newblock Learning transferable visual models from natural language supervision.
\newblock In {\em Proceedings of ICML}, 2021.

\bibitem{hubness}
M.~Radovanovic et~al.
\newblock Hubs in space: Popular nearest neighbors in high-dimensional data.
\newblock {\em Journal of Machine Learning Research}, 11:2487--2531, 2010.

\bibitem{sbert}
N.~Reimers and I.~Gurevych.
\newblock Sentence-{BERT}: Sentence embeddings using siamese {BERT}-networks.
\newblock In {\em Proceedings of EMNLP}, 2019.

\bibitem{pamap2}
A.~Reiss and D.~Stricker.
\newblock Introducing a new benchmarked dataset for activity monitoring.
\newblock In {\em 16th IEEE International Symposium on Wearable Computers (ISWC)}, pages 108--109, 2012.

\bibitem{cadavae}
E.~Schonfeld et~al.
\newblock Generalized zero- and few-shot learning via aligned variational autoencoders.
\newblock In {\em Proceedings of CVPR}, 2019.

\bibitem{smith2017isf}
Samuel~L. Smith, David H.~P. Turban, Steven Hamblin, and Nils~Y. Hammerla.
\newblock Offline bilingual word vectors, orthogonal transformations and the inverted softmax.
\newblock In {\em 5th International Conference on Learning Representations (ICLR 2017)}, 2017.

\bibitem{tong2021}
Catherine Tong, Jinchen Ge, and Nicholas~D. Lane.
\newblock Zero-shot learning for {IMU}-based activity recognition using video embeddings.
\newblock {\em Proceedings of the ACM on Interactive, Mobile, Wearable and Ubiquitous Technologies}, 5(4):1--23, 2021.

\bibitem{xian2018}
Yongqin Xian, Christoph~H. Lampert, Bernt Schiele, and Zeynep Akata.
\newblock Zero-shot learning---{A} comprehensive evaluation of the good, the bad and the ugly.
\newblock {\em IEEE Transactions on Pattern Analysis and Machine Intelligence}, 41(9):2251--2265, 2018.

\bibitem{xu2025}
Lilin Xu, Kaiyuan Hou, and Xiaofan Jiang.
\newblock Exploring the capabilities of {LLM}s for {IMU}-based fine-grained human activity understanding.
\newblock In {\em Proceedings of the 2nd International Workshop on Foundation Models for Cyber-Physical Systems \& Internet of Things (FMSys '25)}, Irvine, CA, USA, 2025. ACM.
\newblock arXiv:2504.02878.

\bibitem{su2025}
Hua Yan, Heng Tan, Yi~Ding, Pengfei Zhou, Vinod Namboodiri, and Yu~Yang.
\newblock Large language model-guided semantic alignment for human activity recognition.
\newblock {\em Proceedings of the ACM on Interactive, Mobile, Wearable and Ubiquitous Technologies}, 8(4), 2025.

\bibitem{unimts}
Xiyuan Zhang, Diyan Teng, Ranak~Roy Chowdhury, Shuheng Li, Dezhi Hong, Rajesh~K. Gupta, and Jingbo Shang.
\newblock {UniMTS}: Unified pre-training for motion time series.
\newblock In {\em Advances in Neural Information Processing Systems}, volume~37, pages 107469--107493, 2024.

\end{thebibliography}

\end{document}